# An Intellectual Property Entity Recognition Method Based on Transformer and Technological Word Information

Yuhui Wang, Junping Du*, Yingxia Shao

( School of Computer Science, Beijing Key Laboratory of Intelligent Telecommunication Software and Multimedia, Beijing University of Posts and Telecommunications, Beijing 100876, China )

**Abstract** Patent texts contain a large amount of entity information. Through named entity recognition, intellectual property entity information containing key information can be extracted from it, helping researchers to understand the patent content faster. Therefore, it is difficult for existing named entity extraction methods to make full use of the semantic information at the word level brought about by professional vocabulary changes. This paper proposes a method for extracting intellectual property entities based on Transformer and technical word information , and provides accurate word vector representation in combination with the BERT language method. In the process of word vector generation, the technical word information extracted by IDCNN is added to improve the understanding of intellectual property entities Representation ability. Finally, the Transformer encoder that introduces relative position encoding is used to learn the deep semantic information of the text from the sequence of word vectors, and realize entity label prediction. Experimental results on public datasets and annotated patent datasets show that the method improves the accuracy of entity recognition.

**Key words** Chinese named entity recognition; intellectual property ; Transformer encoder

CLC number TP391

## 1 Introduction

With the rapid development of science and technology, the speed of technological iteration is also accelerating, and the number of intellectual property resources has exploded. Through patent analysis, valuable information such as the relationship between technologies and the trend of technological development can be revealed. Patent literature contains a large number of specialized vocabulary, and patent analysts with professional background also need to invest a lot of time and cost to understand the content of the patent. Therefore, the automatic extraction of core technical information in patent texts is of great significance to help researchers quickly understand patent information [1][2][3].

Patent documents are compiled by professional technicians, which contain a large number of professional vocabulary and technical terms, and have the characteristics of precise language, complex semantic information, and high information density. Word vector is the core representation technology of natural language processing [4], but traditional word vector representation such as word2vec has many limitations in terms of word polysemy [5]. Bert method [6] can capture information such as word sequence and contextual relationship in the whole sentence, solve the problem of polysemy, and has strong text feature representation ability. At present, the LSTM-CRF method system commonly used in entity recognition is widely used in entity extraction in social media [7] and other scenarios[8]. It can make better use of the global structure information of the text, but it is not sensitive to the information at the word level and cannot make full use of the information in the patent text, semantic information for technical terms. In recent years, the Transformer method [9] has been widely used in natural language processing tasks and achieved good results. The Transformer is based on a self-attention mechanism, which takes into account local and contextual features, can avoid the segmentation of text semantic features, and has excellent features. extraction capacity.

**Fund Project** : This work was supported by National Key R&D Program of China (2018YFB1402600), and the National Natural Science Foundation of China ( 61772083).
**Corresponding author** : Junping Du ( junpingdu @126.com )

The contributions of this paper are as follows:

1）This paper proposes an intellectual property entity recognition method (BWET) based on Transformer and technical word information. By introducing dilated convolutional network, the technical word information in patent text can be extracted, and the ability of word vector to represent intellectual property entities can be improved.

2）This paper proposes to use Transformer to extract deep semantic information, reduce the loss of semantic information, and introduce relative position encoding to overcome the problem of Transformer's lack of relative position awareness and provide excellent feature extraction capabilities.

## 2 Related work

Intellectual property entities refer to words that appear in patent documents containing technical information, mainly including entities that reflect science and technology, method theory, and their fields, such as autonomous driving, lidar, etc. Intellectual property entities may be composed of multiple words, entity boundaries are not easy to distinguish, and each sentence in the patent text contains a large number of entities, which requires more accurate and richer features to describe intellectual property entities. At present, the extraction of intellectual property entities is mainly divided into two categories [10], one is based on traditional machine learning methods[11], and the other is based on deep learning methods[12].

Methods based on traditional machine learning mainly use dependency syntax analysis methods, domain dictionaries and other text analysis methods combined with traditional statistical machine learning methods [13][14] to realize the extraction of intellectual property entities. Chen et al. [15] used the bootstrapping algorithm and added rules to deal with special cases, reduce the impact of semantic drift, and achieve entity extraction in patents with low computational overhead, but the extraction effect on long difficult sentences and complex technical entities composed of multiple words was poor[16].

Currently, more methods are used based on deep learning methods [18][19] to complete sequence labeling tasks and achieve entity extraction. While having better performance, it almost does not require manual construction and selection of features [20]. At present, the most widely used methods mainly include word embedding layer, feature extraction layer and sequence prediction layer. The layer generates the result of sequence labeling, in which sequence prediction mostly uses conditional random field (CRF) to learn the dependencies between label sequences, and optimize the output label sequence. The optimization of the method mainly focuses on the word embedding layer and the sequence prediction layer.

For professional domain entities, the recognition accuracy is improved by improving the word embedding layer and introducing domain professional word features. Wang et al. [21] used the sequence-to-sequence (seq2seq) method to extract the semantic features of patent texts in the communication field. Saad et al. [22] used the BERT (Bidirectional Encoder Representations from Transformers) method to generate dynamic word vectors for biological corpus. Adding the word features of the text to the word vector can alleviate the quality reduction of the word vector caused by the data sparse and OOV problem (out-of-vocabulary) [23]. CNN [24] and Chinese word segmentation dictionary [25] are commonly used word feature construction methods. However, intellectual property entities have a large length span and may be composed of multiple words, and the above methods may lead to the segmentation of word semantics. Yan et al. [26] used the Hidden Markov method [27] to dynamically update the word segmentation dictionary, which improved the word segmentation effect. Feature Extraction Layer The most widely used method structure at present is to [28] Jin et al. [29] proposed to add an attention mechanism layer to further extract the hidden semantic relationship of the text, but the problems that the training speed cannot be accelerated in parallel, and the text information will be lost with recursion are still unsolved. In recent years, as Transformer has been widely used in other tasks of natural language processing and achieved excellent results, improved methods based on Transformer [30] have also been increasingly used in named entity extraction.

## 3 Intellectual property entity recognition

## method based on Transformer and technical word information

This paper proposes an entity extraction method (BWET) based on Transformer and technical word information. Its structure is shown in Figure 1. The method is mainly divided into technical word information fusion layer, intellectual property semantic extraction layer and entity label optimization layer. The technical word information fusion layer generates dynamic word vectors containing contextual semantic information according to the input patent text through BERT and solves the problem of ambiguity of intellectual property entities. The technical word information generated by the word vector after iterative dilated convolutional network (IDCNN) is spliced with the original word vector, and the word vector representation of the fusion technical word information is obtained. The Transformer encoding layer is composed of multiple Transformer encoders, and uses the attention mechanism to extract the deep semantic features of the word vector sequence. The CRF layer is responsible for learning the dependencies between labels and optimizing the output label sequence. The overall flow of the BWET method is shown in Table 1.

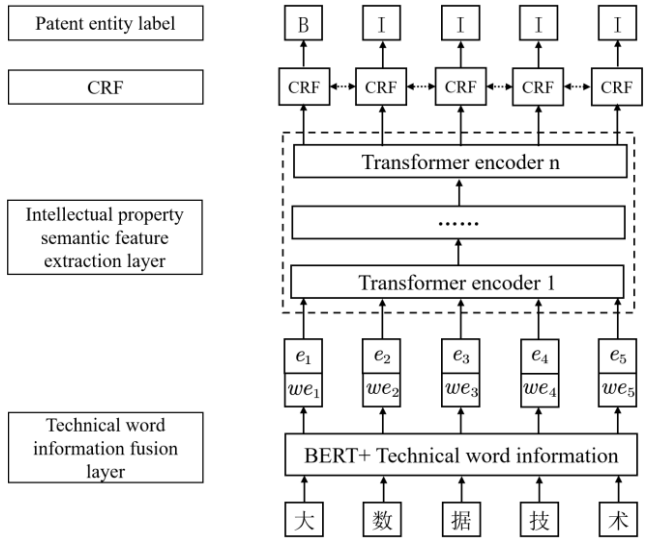

Figure 1 Entity extraction method based on Transformer and technical word information (BWET)

**Table 1 Entity extraction method based on Transformer and technical word information**

| Entity extraction method based on Transformer and technical word information |
| --- |
| Input: Patent text sequence |
| Output: Sequence of IP entity annotations |
| 1) Generate word vectors using BERT on the input text sequence |
| 2) Do the following on word vectors: <br> a) Calculate technical word information using IDCNN <br> b) Concatenate word vectors with technical word information <br> c) Obtain word vector fused with technical word information |
| 3) Use Transformer to extract deep semantic information for word vectors fused with technical word information |
| 4) Generate the probability that each word corresponds to the intellectual property entity label according to the semantic information |
| 5) Use CRF to optimize the entity tag sequence |
| 6) Returns a sequence of IP entity annotations |

### 3.1 Technical word information fusion layer

The function of the technical word information fusion layer is to convert the patent text into a semantic vector representation. Its structure is shown in Figure 2. It is mainly composed of the BERT method and the technical word vector generation method. The final generated word vector is generated by BERT. The initial word vector is concatenated with the word vector.

In addition to complex semantics and high information density, intellectual property entities in patents are also complex in structure and may be composed of many words. It is necessary to combine word information to distinguish intellectual property entities well. Although BERT can solve the problem of ambiguity of text, it cannot provide word features. The traditional word embedding method based on CNN or dictionary lacks flexibility, and in the face of intellectual property entities with variable length and structure, it is easy to cause segmentation of word information.

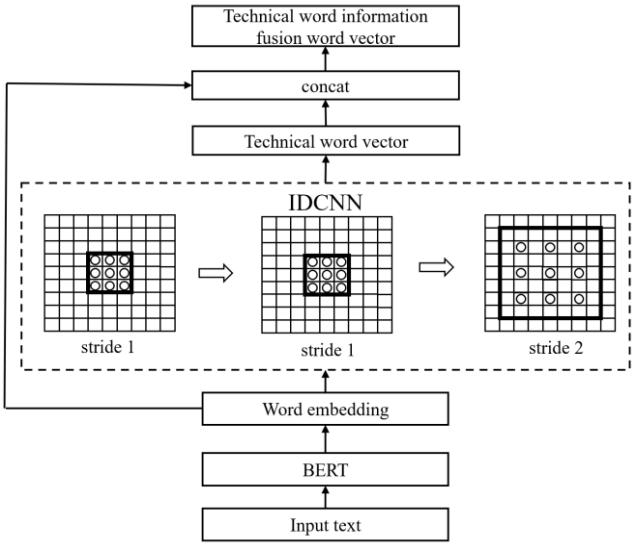

Figure 2 Technical word information fusion layer structure

This paper introduces iterative dilated convolutional network (IDCNN) to realize the generation of technical word vectors in patents. Dilated convolution sets a sequence of dilation steps for the filter, which ignores all input data in the dilation step, while keeping the size of the convolution kernel constant. As the number of convolutional layers increases, the field of view expands exponentially ; as shown in Figure 2, when the step size is 1, the receptive field of view of the expanded convolutional layer is in the bold box $3 \times 3$ Matrix, with a stride of 2, which expands the window of the receptive field of the dilated convolution to a window of $3 \times 3$ and $7 \times 7$. The dilated convolutional network for generating word vectors consists of one dilated convolutional unit, which consists of three dilated convolutional layers with dilation strides 1, 1, and 2. The first two layers of convolutional networks make each feature of the text vector extracted by the dilated convolution , making full use of the effective information of the original text word vector , and the last convolution adopts a larger dilation step size to obtain a wider input matrix data. .

### 3.2 IP semantic extraction layer

The sentences of patent texts are long, the semantic information density is high, and the sentences contain many entities, which are very sensitive to the information loss in the process of semantic extraction. Therefore, the Transformer encoder is introduced to complete the extraction of semantic information, which is mainly composed of a multi-head attention mechanism and a feedforward neural network. The input information is completely transmitted to the output vector through the residual connection, which effectively alleviates the problem of gradient disappearance in the method. However, since the self-attention mechanism does not have a convolutional or recursive network structure, it cannot rely on the network structure to encode the position information of the text sequence in the feature extraction process like CNN or LSTM network, so it is necessary to embed the position vector in the input word vector. The absolute position code is directly added to the word vector, and the position vector is constructed by using sine and cosine codes with different frequencies . The specific calculation process is shown in formula (1) .

$$PE_{(p,2m)} = \sin\left(\frac{p}{10000^{2m/d}}\right)$$
$$PE_{(p,2m+1)} = \cos\left(\frac{p}{10000^{2m/d}}\right) \quad (1)$$

Which $p$ represents the position of the word, $d$ represents the dimension of the position vector, and $m \in [0, d/2 - 1]$ represents a specific dimension of the vector. However, this encoding method will lose the relative position information after passing through the attention mechanism [31], only the absolute position information can be retained, and the relative position is the key feature required for named entity extraction.

According to the self-attention mechanism and the absolute position embedding encoding method, the calculation process of the attention score of the th word and the th word in the sequence is shown in formula $i(2\ j)$ .

$$A_{i,j} = [(e_i + pe_i)W^q][(e_j + pe_j)W^k]$$
$$A_{i,j} = \underbrace{e_i W^q (W^k)^T e_j^T}_{(a)} + \underbrace{pe_i W^q (W^k)^T e_j^T}_{(b)} + \\ \underbrace{e_i W^q (W^k)^T pe_j^T}_{(c)} + \underbrace{pe_i W^q (W^k)^T pe_j^T}_{(d)} \quad (2)$$

where $e$ represents the word vector, $pe$ represents the absolute position embedding vector, $W^q$ and is $W^k$ the weight matrix for generating the query vector and key vector , respectively . Only part of it contains the position information of the two words, but the $(d)$ dot product of the two weight matrices causes the loss of the relative position.

In order to solve this problem, this paper can generate features containing direction information by introducing relative position encoding. The calculation process is shown in formula (3).

$$r_{i,j} = \left[ \ldots \sin\left(\frac{i-j}{10000^{2m/d}}\right) \cos\left(\frac{i-j}{10000^{2m/d}}\right) \ldots \right]^T \quad (3)$$

where $i$ and $j$ represents the position of the input word vector in the text. Using the properties of the sine function $\sin(i-j) = -\sin(j-i)$ to reflect the relative position changes between characters and using the properties of the cosine function $\cos(i-j) = \cos(j-i)$ to reflect the absolute position relationship, both relative position and absolute position information are preserved in relative position encoding. After the relative position is introduced, the query vector only needs to focus on the relative position of the key vector, and the key vector does not need to calculate the position information of the query vector, and the key vector and the relative position vector are retained, so that the attention score can contain more information. The modified attention score calculation formula is shown in Equation (4).

$$A_{i,j} = e_i W^q [(e_j + r_{i,j}) W^k] + u e_j W^k + v r_{i,j} \quad (4)$$

where $e$ represents the word vector, $r_{i,j}$ represents the relative position embedding vector, $W^q$ and $W^k$ are the weight matrices for generating the query vector and the key vector, respectively, and $u$ and $v$ are the learnable parameters.

**3.3 Entity tag optimization layer**

After the patent text is fused with technical word information and Transformer encoder, the probability that each word belongs to each type of label can be calculated according to the semantic information of the text. However, because the dependencies between tags are not considered, for example, the IX tag must appear after the BX tag, which leads to the generation of invalid tag sequences such as no entity start tag, which reduces the entity recognition effect.

In this paper, the conditional random field is introduced to learn the dependencies between labels, which limits the relationship between the generated labels and improves the recognition effect of intellectual property entities. For the label sequence $y = \{y_1, y_2, \ldots, y_n\}$ to be predicted with a length of $n$, the probability that the generated sequence is labeled as a label sequence is obtained by the score function expressed by Equation (5). Use the Viterbi algorithm to find the optimal tag sequence.

$$s(x,y) = \sum_{i=1}^{n} A_{y_{i-1}, y_i} + \sum_{i=1}^{n} P_{i, y_i}$$
$$\log P(y|x) = s(x,y) - \log \sum_{y'} \exp(s(x,y')) \quad (5)$$

where is $s(x,y)$ the score $A_{y_{i-1}, y_i}$ of the label sequence ; $y$ represents the probability of $i-1$ transferring from the $i$ th label to the th label; is $i$ the probability that the $P_{i, y_i}$ th word is mapped to the $i$ th label; $P(y|x)$ is the probability that the generated sequence is labeled as a label sequence .

## 4 Experimental results and analysis

**4.1 Experimental data and evaluation indicators**

The experiment uses the accuracy rate ( precision , P ), recall rate (recall , R ) and F1 value (F1-Score , F 1 ) as the evaluation indicators of the BERT-based fusion feature text classification method.

The public dataset uses the CLUENER dataset [32], and the training set has 10,748 sentences, including 10 types of entities such as addresses, companies, and movie names. The granularity of entity categories in this dataset is relatively fine, and there are entities with similar semantic information such as book titles and movie titles, which are closer to the scene of intellectual property entity recognition. The patent entity recognition data set has a total of 1972 patent titles and abstracts. The entities in them are manually marked and divided into training set and test set according to the ratio of 4: 1. The distribution of entity types is shown in Table 2.

Experimental data are labeled with BIO : label each element as "BX" , "IX" or "O" . Among them, "BX" indicates that the fragment where this element is located is of type X and this element is at the beginning of this fragment, "IX" indicates that the fragment where this element is located is of type X and this element is in the middle of this fragment, "O" means no of any type. For example, if X is represented as a technical phrase (TECH), the three BIO tokens are: B-TECH : the beginning of a technical phrase, I-TECH : the middle part of a technical phrase, O : a character that is not a technical phrase.

Table 2 Types and distribution of intellectual property entities

| serial number | entity type | entity meaning | number of entities |
|---|---|---|---|
| 1 | DOM | Field of study | 4 925 |
| 2 | TECH | technical terms | 2 6027 |
| 3 | USED | Application direction | 4 393 |
| 4 | EFF | Efficacy word | 5594 |
| 5 | INFO | data source | 6659 |
| 6 | MAT | Material | 3289 |

**4.2 Experimental parameters**

In the experiment, the BERT-wwm-ext trained by the Xunfei joint laboratory of Harbin Institute of Technology is used . There are 12 layers of Transformer structure, 12 attention heads, the generated word vector dimension is 768 dimensions, and the method does not participate in parameter adjustment ( No fine-tuning of BERT).

The convolution kernel size of DCNN is set to 3, the number of convolution kernels is set to 128, the dilation step size is set to (1, 1, 2 ), the number of layers of Transformer is 4 layers, and there are 16 attention heads. The maximum length of a single sentence is 128, the size of batch_size is set to 32, the size of epoch is set to 100 , the learning rate is set to 0.0001 , the dropout is set to 0.5 during training, and the optimizer is Adam. In the comparison method, the number of hidden layer neurons of BiLSTM is set to 256, the number of IDCNN convolution kernels is set to 128 , and the learning rate is set to 0.0 01. The rest of the training parameters are the same as those of the BWET method.

**4.3 Experimental results**

1) Experiment 1: Validation of the BWET method on public datasets

In order to verify the effectiveness of the method proposed in this paper, the BiLSTM-CRF, IDCNN-CRF and Transformer-CRF methods with relative positions are used to generate word vectors using word 2 Vec and BERT, respectively, and comparative experiments are carried out on public datasets.

Table 3. Comparison of entity recognition effects of each method on the CLUENER dataset

| method | Precision _ _ | Recall | F1 _ |
|---|---|---|---|
| BiLSTM-CRF | 0.7275 _ | 0.6867 _ | 0.7065 _ |
| IDCNN-CRF | 0.6439 _ | 0.6776 _ | 0.6603 _ |
| Transformer-CRF | 0.7278 _ | 0.7214 _ | 0.7245 |
| BERT | 0.7845 | 0.7666 _ | 0.7754 |
| BERT-BILSTM-CRF | 0.7857 | 0.8210 _ | 0.8029 |
| BERT-IDCNN-CRF | 0.7958 _ | 0.7585 _ | 0. 7767 |
| - Transformer-CRF | 0.7863 | 0.8350 _ | 0.8099 |
| **BWET** | **0.7912 _** | **0.8473 _** | **0.8183 _** |

can be seen from Table 3 , the F 1 value of the proposed BEWT method reaches 0.8183 , which achieves the best performance and has a huge improvement in entity recognition performance. Compared with other methods, the performance of the IDCNN method has a big gap, indicating that the weak context-dependent feature acquisition ability of the IDCNN network has greatly restricted entities when dealing with long texts. The accuracy of the BERT-IDCNN method is higher than other methods, indicating that word features can improve the accuracy of entity recognition. Compared with the BERT recall rate and F1 value, the BERT-BiLSTM method has improved by 5 % and 3%, indicating that the LSTM method can learn the contextual semantic information of the text well, but the loss of information in the transmission process affects its performance. . The Transformer method significantly reduces the information loss in the feature extraction process through the attention mechanism and residual connection. Compared with the BiLSTM method, the recall rate is improved by 4%, and it is also improved by 1% after the introduction of BERT.

2) Experiment 2: Validation of the BWET method on the patent dataset

In order to verify that the BWET method has a stronger entity recognition effect for patent data, the same comparison method as Experiment 1 is used to conduct a comparative experiment on the patent data set.

**Table 4. Comparison of entity recognition effects of various methods on patent datasets**

| method | Precision | Recall | F1 _ |
|---|---|---|---|
| BiLSTM-CRF | 0.5914 _ | 0.5166 _ | 0.5515 _ |
| IDCNN-CRF | 0.4607 _ | 0.4378 _ | 0.4489 _ |
| Transformer-CRF | 0.5894 | 0.5747 | 0.5819 |
| BERT | 0.7089 | 0.6971 _ _ | 0.7029 |
| BERT-BILSTM-CRF | 0.7442 | 0.7303 _ | 0.7372 |
| BERT-IDCNN-CRF | 0.6440 | 0.6494 | 0.6467 |
| - Transformer-CRF | 0.7692 | 0.7884 _ | 0.7787 |
| **BWET** | **0.7817 _** | **0.8174 _** | **0.7992 _** |

from Table 4 , in the patent entity identification data set, the technical entities in the patent belong to different entity types because of the different contexts in which they are located. domain, while denoting technical terms in "using artificial intelligence techniques" makes traditional static word vectors perform poorly. After the introduction of BERT, the recognition accuracy rate, recall rate and F 1 value are all improved by 15% , which solves the problem of ambiguity. Since the number and complexity of entities contained in patent texts are higher than general texts, the defects of BiLSTM method's lack of perception of local features and information loss are more obvious on patent datasets. Compared with BERT-BiLSTM, the BERT-Transformer method improves the recognition accuracy by 2.5 % and the recall rate by 5 % compared with the BERT-BiLSTM on the patent dataset , showing a larger performance difference than the general dataset. The BWET method proposed in this paper is more sensitive to the word-level semantic changes brought by technical terms due to the addition of word features to the text embedding, and the F 1 value is increased by 2 % compared with the BERT-Transformer method.

3) Experiment 3: BWET method F1 value changes with the number of training rounds

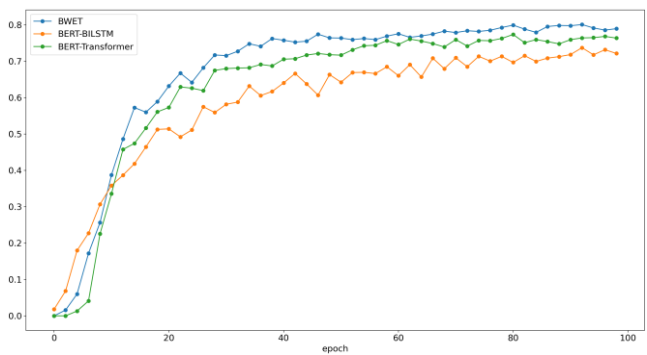

Fig. 3 Change trend of method F 1 value during training

Figure 3 compares the changes in the F1 value of the three methods of BWET, BERT-BiLSTM and BERT-Transformer during the training process. As can be seen from Figure 3, at the beginning of training, the F1 value of BERT-BiLSTM is much faster than the Transformer-based method, mainly due to the fact that LSTM has fewer parameters. When the training is about 10 rounds, the better feature extraction ability of Transformer is fully reflected, and the F1 value quickly exceeds that of BERT-BiLSTM, and BWET has improved F1 earlier than the BERT-Transformer method due to the integration of technical word information . , and the increase is higher. In the middle of the training process, the F1 value of the BERT-BiLSTM method often fluctuates greatly. The reason is likely to be the loss of information in the recursive transfer process, while the BWET method can retain more abundantly because of the combined effect of word information and Transformer. semantic information, thereby reducing fluctuations.

4) Experiment 4: Comparison of the recognition effects of different entities in the BWET method

This paper also explores the recognition ability of BWET for each category of entities. Figure 4 shows the accuracy, recall and F 1 value of BWET method for 6 kinds of intellectual property entity recognition.

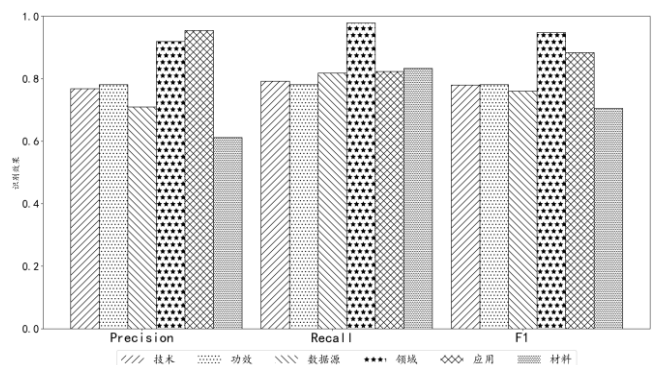

Figure 4 Recognition performance of various types of entities in BWET

Figure 4 , the entity recognition effect of the domain and application direction is significantly better than that of other types of entities. The positions and sentence structures of these two types of entities appearing in the patent text are relatively fixed. For example, the fields they belong to are mainly similar: they belong to the field of intelligent transportation, involving the field of automobile safety; the application

direction is often used for fully automatic driving, based on deep learning Trajectory recognition methods and other forms appear, which are relatively weakly dependent on context. Although the technical entity does not have a fixed sentence structure, and the context page of the sentence in which it is located is relatively complex, it relies on the additional word features to improve the discrimination of the semantic information of the technical entity, and obtains an effect similar to that of the relatively stable functional word form. Material entities are affected by the imbalance of training data, and their recognition effect is poor.

5) Experiment 5: Influence of technical word information dimension on BWET method

This paper studies the influence of the technical word information dimension on the entity recognition effect. Figure 5 shows the recognition effect of the BWET method on the patent data set when the number of convolution kernels of IDCNN is 64, 128 , 256 and 512 respectively.

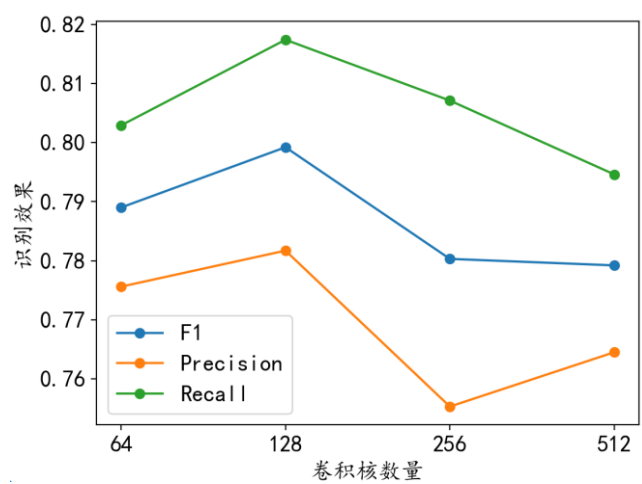

Figure 5 Recognition performance of various types of entities in BWET

As can be seen from Figure 5, when the number of convolution kernels is 64 and 128, the F1 value of the BWET method can reach 0.79 . When the number of convolution kernels exceeds 128, the F1 value is only 0.78 , and the recall rate is only 0.78. There has been a marked decline. The difference in recognition performance caused by the change in the number of IDCNN convolution kernels is close to 2%, indicating that the method is sensitive to this parameter. When the size of the convolution kernel is small, the local semantic features of the text cannot be fully mined, but due to the high density of word information, a better recognition effect can still be achieved. When the number of convolution kernels is large, although the semantic information will not be missed, the semantic information contained in the word vector is more sparse, which weakens the role of the word vector in entity recognition, resulting in a reduction in the recognition effect of the method. In addition, it can be seen that when the quality of technical word information decreases, the recall rate drops significantly and the accuracy rate fluctuates to a certain extent, indicating that the main function of technical word information is to improve the recall rate of the method.

## 5 Conclusion

In view of the fact that intellectual property entities in patent texts have strong context dependencies, and the composition of semantic information and entity words is relatively complex, it is difficult for existing methods to utilize semantic information at the level of intellectual property entity words.

intellectual property entity recognition method based on Transformer and technical word information . The method utilizes the dynamic word vector representation provided by the BERT language method, integrates the context information of the patent into the word vector, and solves the problem of ambiguity of entities. The technical word information is extracted by IDCNN and added to the word vector to enhance the perception of term entities. Using the Transformer encoder that introduces relative position encoding, the Transformer encoder can perceive the relative position and front-to-back direction of the text sequence, and improve the ability to extract semantic information. Experiments show that the method proposed in this paper can effectively improve the performance of named entity recognition on general corpus and patent datasets.

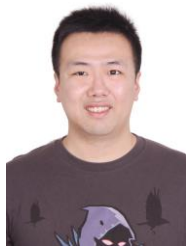

**Yuhui Wang**, (1996-), male, master, member of CCF. The main research areas are natural language processing and data mining

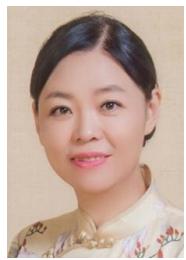

**Junping Du**, (1963-), female, professor, CCF Fellow (NO.10411D). Main research areas are artificial intelligence, machine learning and pattern recognition

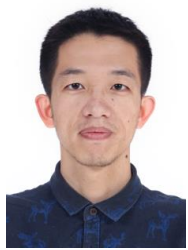

**Yingxia Shao**, (1988-), male, associate professor, senior member of CCF. The main research areas are large-scale graph analysis, parallel computing framework and knowledge graph analysis